\documentclass[times,review,10pt]{elsarticle}

\usepackage[numbers]{natbib}
\usepackage{soul}
\usepackage{url}
\usepackage{bm}
\usepackage{wrapfig}
\usepackage{algorithm}
\usepackage{graphicx}
\usepackage{subcaption}
\usepackage{caption}
\usepackage{amsmath}
\usepackage{amsfonts}
\usepackage{amssymb}
\usepackage{todonotes}
\usepackage{algpseudocode}
\usepackage{booktabs,multirow}
\usepackage{pifont}
\usepackage{siunitx}
\usepackage{cleveref}
\usepackage{todonotes}
\newcommand{\cmark}{\ding{51}}%
\newcommand{\xmark}{\ding{55}}%

\def\tsc#1{\csdef{#1}{\textsc{\lowercase{#1}}\xspace}}
\tsc{WGM}
\tsc{QE}
\tsc{EP}
\tsc{PMS}
\tsc{BEC}
\tsc{DE}

\begin{document}

\begin{frontmatter}

\title{Pool-Select-Refine for Allocation-Aware Generative Dataset Distillation}

\author[ufukui]{Wenmin Li}
\ead{l26008w@u-fukui.ac.jp}

\author[swun]{Zhongkai Zhao}

\author[ufukui]{Shunsuke Sakai\corref{cor1}}

\author[ufukui]{Tatsuhito Hasegawa}

\cortext[cor1]{Corresponding author}

\affiliation[ufukui]{
organization={Graduate School of Engineering, University of Fukui},
addressline={3-9-1 Bunkyo},
city={Fukui},
postcode={910-8507},
state={Fukui},
country={Japan}
}

\affiliation[swun]{
organization={College of Computer Science and Artificial Intelligence, Southwest Minzu University},
addressline={No.168, Wenxing Section, Dajian Road, Konggang Development Zone, Shuangliu District},
city={Chengdu},
postcode={610225},
state={Sichuan},
country={China}
}

\begin{abstract}
Diffusion-based dataset distillation has recently emerged as a promising paradigm for condensing large-scale datasets into compact synthetic sets. By leveraging pretrained generative priors, these methods can produce realistic class-conditional samples more efficiently than traditional matching-based approaches. However, most existing diffusion-based methods still adopt a rigid ``\textit{Generate-and-Use}'' strategy, where the generated samples are directly treated as the final distilled set under a fixed images-per-class budget. Such a design tightly couples candidate generation with final budget allocation, which may result in redundant waste of the limited budget or insufficiently informative samples.
In this paper, we propose ``\textit{Pool-Select-Refine}'', a two-stage framework for allocation-aware generative dataset distillation. First, instead of directly using a fixed number of generated samples, we construct an over-complete candidate pool and select a compact subset under the target budget. Second, we refine the selected samples in latent space using soft-label supervision derived from the teacher model, improving semantic alignment while preserving the generative prior. This design explicitly decouples generation, selection, and refinement, enabling more effective use of the distillation budget.
Experiments on large-scale and fine-grained image classification benchmarks show that the proposed framework delivers consistent gains over diffusion-based baselines. The results suggest that introducing a curation stage before refinement is a simple yet effective way to improve diffusion-based dataset distillation.
\end{abstract}


\end{frontmatter}

\section{Introduction}
\label{sec:introduction}

The performance of modern pattern-recognition systems often depends on large annotated datasets, yet storing, transferring, and repeatedly training on such datasets can be costly under limited memory, communication, or evaluation budgets. Dataset Distillation (DD) \cite{wang2020datasetdistillation} addresses this problem by synthesizing a compact training set that preserves the utility of the original data. Recent surveys further position DD as a data-efficient learning paradigm connecting dataset compression, training efficiency, and compact data representation \cite{10.1109/TPAMI.2023.3323376}. From the perspective of Pattern Recognition, this goal is not only to reduce training cost, but also to construct compact data representations that remain useful under practical recognition conditions. Recent studies have therefore extended DD toward trustworthy recognition \cite{MA2025110875}, point-cloud condensation \cite{ZHANG2026112494}, and reliable evaluation under distribution shifts \cite{ZHANG2025110926}, all of which suggest that distilled data should provide reliable and informative supervision rather than merely visually plausible samples.

\begin{figure}[t]
    \centering
    \includegraphics[width=0.82\linewidth]{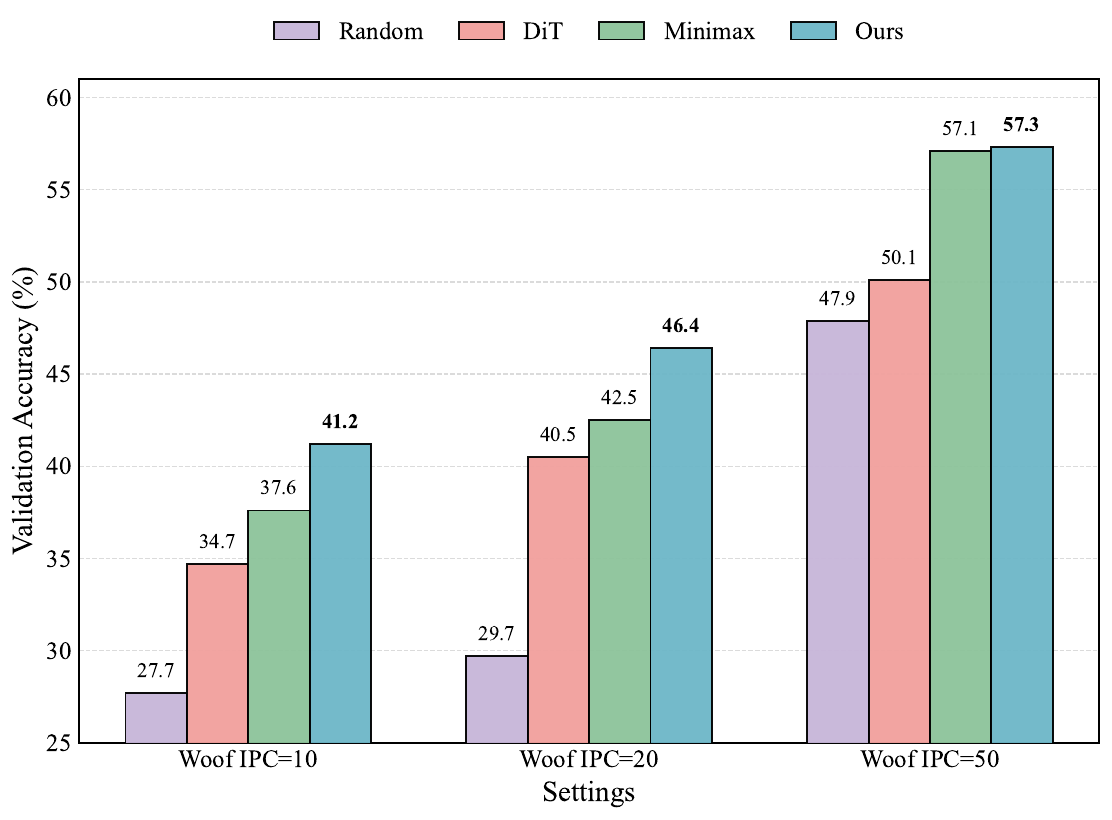}
    \caption{\textbf{Performance comparison on ImageWoof under different IPC budgets.} 
    Our method consistently improves over the baselines under the same IPC budget, demonstrating the effectiveness of explicit curation and refinement for fine-grained dataset distillation.}
    \vspace{-6pt}
    \label{fig:fig1}
\end{figure}

Existing DD methods can be roughly divided into optimization based and generative paradigms. Kernel-based methods, such as KIP \cite{50025}, FRePo \cite{Zhou2022FRePo}, and random-feature approximation \cite{Loo2022RFAD}, formulate distillation through predictive or feature-level matching. Gradient- and distribution-matching methods optimize synthetic images by matching gradients or feature distributions between real and synthetic data \cite{DBLP:journals/corr/abs-2006-05929,Zhao2021DSA,690345d30c9f4e6dbeef4ad9210bee62}, with later improvements based on feature alignment \cite{Wang2022CAFE}, attention matching \cite{Sajedi2023DataDAM}, and mutual-information maximization \cite{Shang2023MIM4DD}. Another line of work matches training dynamics, including MTT \cite{cazenavette2022distillation}, DATM \cite{Guo2024DATM}, and FTD \cite{Du2023FTD}. Although effective, these methods may become expensive when scaling to high-resolution or fine-grained recognition settings.

Generative DD has emerged as a scalable alternative by exploiting pretrained generative priors. Early studies use GAN priors to synthesize informative samples \cite{zhao2022synthesizing}, while GLaD introduces deep generative priors \cite{cazenavette2023glad} and later work explores hierarchical GAN features \cite{DBLP:conf/cvpr/ZhongFCGQQX25}. More recently, diffusion models \cite{Ho2020DDPM,Song2021DDIM} have become attractive because of their stable generation and broad mode coverage compared with traditional GANs \cite{10.1145/3422622}. Along this direction, Minimax Diffusion formulates generation as an importance-aware minimax process \cite{Gu_2024_CVPR}, D4M studies disentangled diffusion modeling \cite{DBLP:conf/cvpr/SuHG0T24}, and IGD introduces influence-guided diffusion \cite{chen2025influenceguided}. Other recent variants further investigate mode-guided generation \cite{santiago2025mgd}, practical diffusion-based unlocking strategies \cite{moser2026unlocking}, information-guided sampling \cite{ye2025informationguided}, adversary-guided curriculum sampling \cite{zou2025enhancing}, and consistency rectification \cite{11444255}.


\begin{table*}[t]
\centering
\setlength{\tabcolsep}{6pt}
\renewcommand{\arraystretch}{1.25}
\caption{\textbf{Mechanism-level comparison of representative DD methods.} ``\cmark'' indicates explicit support, and ``\xmark'' indicates no support.}
\label{tab:gen_dd_comparison}
\resizebox{\textwidth}{!}{
\begin{tabular}{lccccc}
\toprule
Method & Generative Prior & Explicit Curation Stage & Budget Allocation &
Teacher Soft Targets Reused & Teacher-guided Latent Refinement\\
\midrule
DM \cite{690345d30c9f4e6dbeef4ad9210bee62} & \xmark & \xmark & \xmark & \xmark & \xmark \\
MTT \cite{cazenavette2022distillation} & \xmark & \xmark & \xmark & \xmark & \xmark \\
GLaD \cite{cazenavette2023glad} & \cmark & \xmark & \xmark & \xmark & \xmark \\
Minimax \cite{Gu_2024_CVPR} & \cmark & \xmark & \xmark & \xmark & \xmark \\
IGD \cite{chen2025influenceguided} & \cmark & \xmark & \xmark & \xmark & \xmark \\
\midrule
Ours & \cmark & \cmark & \cmark & \cmark & \cmark \\
\bottomrule
\end{tabular}
}
\end{table*}

However, improving the generation process alone does not guarantee efficient use of the final IPC (Image Per Class) budget. In a typical ``\textit{Generate-and-Use}'' pipeline, a generator such as DiT \cite{Peebles2023DiT} produces a fixed number of samples per class, and all generated samples are directly treated as the distilled set. Because diffusion sampling is stochastic, this fixed set may contain redundant modes, weakly informative instances, or semantically unstable candidates. As summarized in Table~\ref{tab:gen_dd_comparison}, existing generative methods exploit strong priors, but they usually do not explicitly insert a curation stage between candidate generation and final budget allocation. We refer to this problem as \textbf{allocation waste}: scarce IPC slots may be spent before verifying whether the generated samples are useful for student training.

To address this issue, we propose ``\textit{Pool-Select-Refine}'', a two-stage framework for allocation-aware generative dataset distillation. Instead of directly accepting the first $K$ generated samples per class, Stage I constructs an over-complete class-conditional candidate pool and selects a compact subset according to semantic reliability, feature-space diversity, and predictive uncertainty. Stage II further refines the selected samples in latent space using teacher-derived soft-label supervision while preserving the pretrained generative prior \cite{qin2024a}. Only the cached latent codes of selected samples are optimized, making the framework lightweight, generator-agnostic, and suitable for fine-grained recognition scenarios where both semantic fidelity and budget efficiency matter.

\begin{figure}[t]
    \centering
    \includegraphics[width=1.0\linewidth]{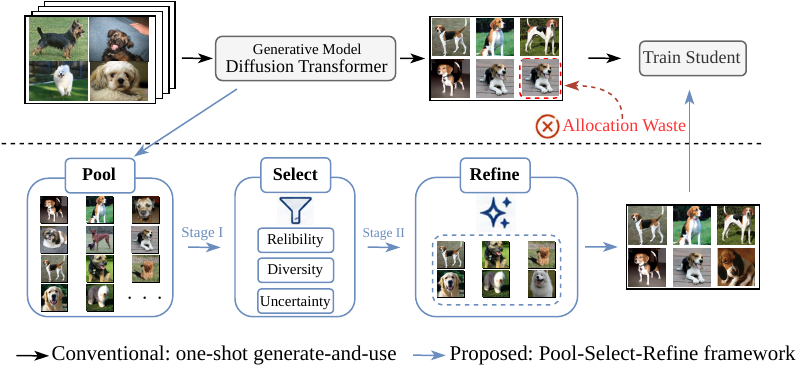}
    \caption{\textbf{Pipeline comparison between conventional generate-and-use distillation and our Pool-Select-Refine framework.} Our method first constructs an over-complete candidate pool, selects useful samples, and then refines the selected subset before student training.}
    \label{fig:method10}
\end{figure}

Figure~\ref{fig:method10} illustrates the pipeline-level difference between conventional one-shot generation and our framework. The conventional pipeline directly trains the student on a fixed generated set, whereas our method treats the generator as a candidate source and makes the limited budget more controllable through selection and refinement. As shown in Figure~\ref{fig:fig1}, this design consistently improves over random real-data subsampling and diffusion-based generation baselines on ImageWoof, especially under low-budget settings where each IPC slot is valuable. Extensive experiments further show consistent gains on ImageNet subsets and fine-grained benchmarks while remaining competitive on CIFAR-scale datasets.

The key contributions of this work are as follows:
\begin{enumerate}
    \item We point out that existing diffusion-based DD methods suffer from \textbf{allocation waste} due to the rigid ``\textit{Generate-and-Use}'' paradigm, where limited IPC budgets are directly spent on synthetic samples without an explicit curation stage.
    \item We propose a two-stage ``\textit{Pool-Select-Refine}'' framework to address this issue, which first performs IPC-constrained pool selection and then conducts soft-label-guided latent refinement on the selected samples.
    \item Our framework is generic and easy to integrate with diffusion-based distillation pipelines, delivering consistent improvements on ImageNet subsets and fine-grained recognition benchmarks while remaining competitive on CIFAR-scale datasets.
\end{enumerate}

\section{Preliminary}
\label{sec:preliminary}

\subsection{Generative DD}
\label{subsec:generative distillation}
Generative dataset distillation uses a pretrained generator as the source of synthetic training samples, instead of directly optimizing every synthetic image from scratch. This paradigm is attractive for high-resolution and fine-grained recognition settings because the generator provides a strong image prior and can produce visually realistic class-conditional candidates. IT-GAN introduces a GAN prior to synthesize informative training samples for dataset distillation \cite{zhao2022synthesizing}. GLaD improves the generalization of distilled data by exploiting a deep generative prior \cite{cazenavette2023glad}. More recently, diffusion models have become attractive for generative distillation because they offer stable generation and broad mode coverage. DiT provides a transformer-based diffusion backbone that can generate high-quality class-conditional images \cite{Peebles2023DiT}. Minimax Diffusion formulates diffusion-based dataset distillation as an importance-aware minimax generation process \cite{Gu_2024_CVPR}. D4M studies disentangled diffusion modeling for dataset distillation \cite{DBLP:conf/cvpr/SuHG0T24}. IGD improves sample informativeness by guiding the diffusion process with influence estimation \cite{chen2025influenceguided}.

Despite these advances, most diffusion-based pipelines still follow a fixed ``\textit{Generate-and-Use}'' strategy: the generator produces exactly the target number of samples, and all generated samples are directly used as the distilled set. This strategy leaves no explicit mechanism for deciding whether each generated sample deserves to occupy a limited IPC slot. Our work therefore treats generative distillation as a budget-allocation problem and introduces explicit curation before latent refinement.

\subsection{Teacher-Derived Signals for IPC-Constrained Curation}
\label{subsec:selection baseline}
Given a class $c$ and a target budget of $K$ IPC, assume that a generator first produces an over-complete class-conditional candidate pool $\mathcal{P}_c$, where $|\mathcal{P}_c|>K$. The curation problem is to select a compact subset $\mathcal{S}_c \subset \mathcal{P}_c$ with $|\mathcal{S}_c|=K$. In this paper, we use a fixed teacher model $T$ to provide three types of selection signals: semantic reliability, feature-space coverage, and predictive uncertainty. \\
\textbf{Semantic reliability.}
A natural way to estimate whether a generated candidate matches its target class is to use the teacher's class confidence. Maximum softmax probability is a common confidence signal for recognition and out-of-distribution detection \cite{hendrycks2016baseline}. To obtain a smoother probability distribution, we use temperature scaling, which is widely used for probability calibration \cite{pmlr-v70-guo17a}. The reliability score of a candidate $x$ for class $c$ is defined as
\begin{equation}
    s_{rel}(x) = q_{T}^\tau(c \mid x),
    \label{eq:conf}
\end{equation}
where $q_T^\tau(c \mid x)$ denotes the temperature-scaled teacher probability assigned to class $c$. A larger value indicates stronger semantic consistency with the target class. \\
\textbf{Feature-space coverage.}
Reliability alone may select only easy and prototypical samples. To reduce redundancy, we encourage the selected subset to cover different regions of the teacher feature space. This idea is related to core-set selection, where K-Center selects representative samples by minimizing the distance from each data point to its nearest selected center \cite{sener2018active}. Let $e_T(x)$ denote the teacher feature embedding of $x$. A coverage-oriented subset can be written as
\begin{equation}
\mathcal{S}_c^\star=\underset{\mathcal{S} \subset \mathcal{P}_c,\ |\mathcal{S}|=K}{\operatorname{argmin}} \sum_{x \in \mathcal{P}_c}\min_{x' \in S}\left\| e_T(x) - e_T(x') \right\|_2^2.
\label{eq:cluster_select}
\end{equation}
This objective encourages selected samples to represent the candidate pool more evenly, but it does not explicitly prevent semantically ambiguous samples from being selected. \\
\textbf{Predictive uncertainty.}
Uncertainty is often used in active learning to identify informative samples for annotation \cite{settles2009active}. In our setting, however, excessive uncertainty is undesirable because the candidates are synthetic and the IPC budget is fixed. Therefore, we use predictive entropy as a penalty rather than a preference. This helps avoid allocating scarce budget slots to candidates whose teacher predictions are unstable or ambiguous.
These three signals play complementary roles. Reliability suppresses semantic drift, coverage reduces redundancy, and uncertainty penalizes ambiguous synthetic candidates. Section~\ref{sample selection} combines them into a unified IPC-constrained selection rule.

\subsection{Latent Diffusion Models}
\label{subsec:LDM}
LDMs (Latent Diffusion Models) \cite{Rombach2022LDM} perform the generative process in the compressed latent space of a pre-trained VAE (Variational Autoencoder). Let $\mathcal{E}(\cdot)$ and $\mathcal{D}(\cdot)$ be the VAE encoder and decoder, respectively. An image $x$ is first mapped to a latent code $z_0 = \mathcal{E}(x)$. The diffusion process adds Gaussian noise to $z_0$ over $T$ timesteps, producing a sequence $z_1,...,z_T$. For a generation, a denoising network $\epsilon_\theta(z_t, t, y)$ is trained to predict the added noise $\epsilon$ conditioned on the time step $t$ and class label $y$. The standard training objective is to minimize the simple mean-squared error:
\begin{equation}
  \mathcal{L}_{\mathrm{LDM}}
  = \mathbb{E}_{x,\, \epsilon \sim \mathcal{N}(\mathbf{0},\mathbf{I}),\, t}
  \bigl[ \lVert \epsilon - \epsilon_{\theta}(z_t, t, y) \rVert_2^2 \bigr],
  \label{eq:ldm_loss}
\end{equation}
where $z_t = \sqrt{\bar{\alpha}_t}\, z_0 + \sqrt{1 - \bar{\alpha}_t}\, \epsilon_t
$ and $\bar{\alpha}_t$ controls the noise schedule. In our framework, this denoising objective is later reused during latent refinement as an on-manifold regularizer, encouraging the optimized latent codes to remain consistent with the pretrained diffusion prior.

\begin{figure*}[t]
    \centering
    \makebox[\textwidth][c]{%
    \includegraphics[width=1.08\textwidth]{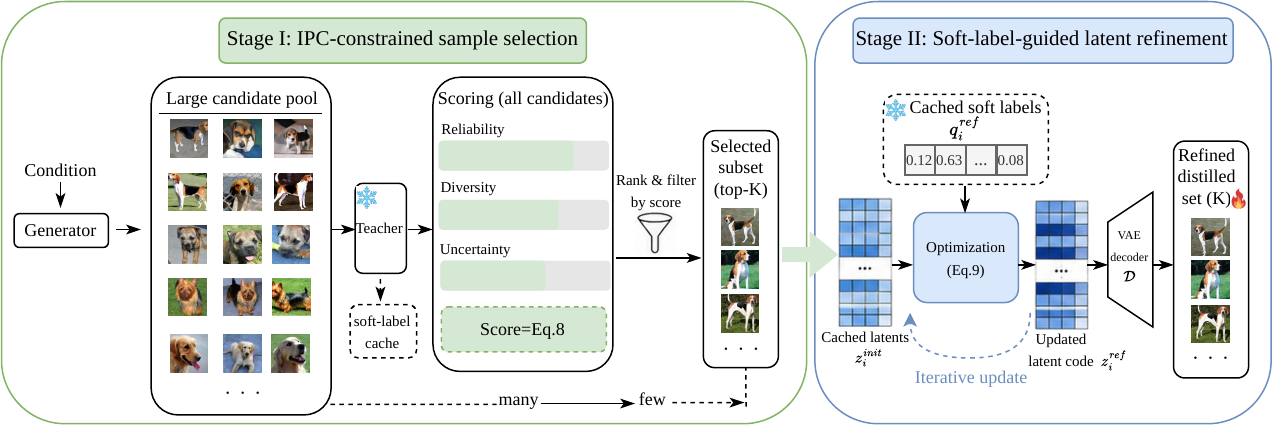}
    }
    \caption{\textbf{Detailed workflow of the proposed Pool-Select-Refine framework.} Stage I constructs an over-complete candidate pool, evaluates all candidates using teacher-derived reliability, diversity, and uncertainty, and selects a compact subset under the target IPC budget. The teacher soft labels and generation latents of the selected samples are cached. Stage II directly optimizes the cached latent codes under soft-label guidance and decodes the refined latents into the final distilled set for student training.}
    \label{fig:intro1}
    \vspace{-4pt}
\end{figure*}

\section{Proposed Method}
\label{approach}
As illustrated in Figure~\ref{fig:intro1}, the proposed ``\textit{Pool-Select-Refine}'' framework consists of two consecutive stages: IPC-constrained pool selection and soft-label-guided latent refinement. Given a target IPC budget, Stage I first generates an over-complete class-conditional candidate pool and selects a compact subset according to teacher-derived reliability, feature-space diversity, and predictive uncertainty. For the selected samples, we cache both their teacher soft labels and generation latents. Stage II then directly refines the cached latent codes using the cached teacher soft labels while freezing the weights of the pretrained generative prior. The final refined samples, together with their teacher soft targets, are used to train the student model.

\subsection{Motivation and Problem Formulation}
\label{subsec:problem definition}
We consider DD with $C$ classes and a fixed budget of 
$K$ IPC. Let $T$ denote a fixed teacher model and 
$S$ a student model to be trained on the distilled set. The goal is to construct a compact distilled dataset:
\begin{equation}
\mathcal{D}^\star=\bigcup_{c=1}^{C}\{(x_{c,k}, q_{c,k}^{\mathrm{ref}})\}_{k=1}^{K},
\label{eq:distilled_set}
\end{equation}
where $x_{c,k}$ is the final distilled sample for class $c$, and
$q_{c,k}^{\mathrm{ref}}$ denotes its associated teacher soft target. Unlike rigid ``\textit{Generate-and-Use}'' pipelines that directly treat exactly $K$ generated samples as the final distilled dataset, we start from an over-complete class-conditional candidate pool. Specifically, for each class $c$, a pretrained generator $G$ produces:
\begin{equation}
    \mathcal{P}_c=\{(z_j^{\mathrm{init}}, x_j^{\mathrm{init}})\}_{j=1}^{M}, M > K, 
    \label{eq:image_pool}
\end{equation}
where $z_j^{\mathrm{init}}$ is the latent code and $x_j^{\mathrm{init}} = Dec(z_j^{\mathrm{init}})$ is the decoded candidate image. This reformulation turns diffusion-based distillation from direct generation into a budget allocation problem: \textit{given a limited budget $K$, how can we assign the final slots to the most useful candidates in $\mathcal{P}_c$}?

\begin{figure}[t]
    \centering
    \includegraphics[width=1.0\linewidth]{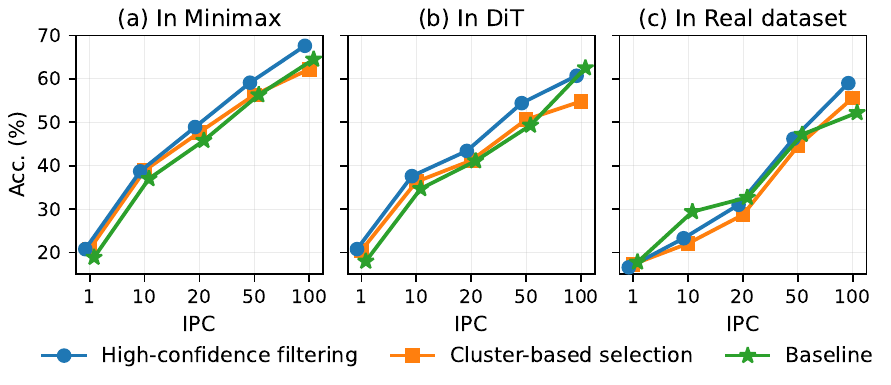}
    \caption{\textbf{Effect of pool-based selection.} Selecting $K$ samples from an over-complete candidate pool improves student accuracy compared with directly using a fixed number of generated samples, indicating that budget allocation is a key factor in diffusion-based dataset distillation. Markers are slightly horizontally shifted for better readability.}
    \label{fig:intro2}
\end{figure}

Figure~\ref{fig:intro2} provides empirical motivation for this formulation. Compared with directly using a fixed number of generated samples, selecting $K$ samples from an over-complete pool consistently improves student performance. This indicates that budget waste is not only a generation-quality issue, but also a budget-allocation issue. Therefore, Stage I focuses on selecting more useful seeds from the candidate pool. Since each selected seed retains its cached latent $z_i^{\mathrm{init}}$ and teacher soft target $q_i^{\mathrm{ref}}$, Stage II can further refine the selected samples directly in latent space without re-encoding the images.

\subsection{IPC-constrained pool selection}
\label{sample selection}
Stage I aims to allocate the fixed IPC budget to the most useful candidates in the over-complete pool. For each class $c$, let
$\mathcal{P}c={(z_i^{\mathrm{init}},x_i^{\mathrm{init}})}{i=1}^{M}$ denote the candidate pool generated under the class condition $c$, where $M>K$. We first query the fixed teacher $T$ for each candidate and obtain its temperature-scaled prediction
$q_i=q_T^\tau(\cdot \mid x_i^{\mathrm{init}})$ and feature embedding
$e_i=e_T(x_i^{\mathrm{init}})$. Candidates whose teacher-predicted label is inconsistent with the target class are filtered out:
\begin{equation}
\mathcal{I}_c={i \mid \arg\max_k q_i(k)=c}.
\end{equation}
If fewer than $K$ candidates remain after filtering, we relax this constraint and keep the top-$K$ candidates according to $q_i(c)$, so that the final IPC budget is always satisfied.

\noindent
\textbf{Reliability ($s_{\mathrm{rel}}$).}
Following the reliability signal defined as Equation~\ref{eq:conf}, we evaluate whether a candidate is semantically consistent with its target class. Candidates whose teacher-predicted label disagrees with $c$ are discarded:
\[
\arg\max_k q_T^\tau(k \mid x) \neq c.
\]
This avoids allocating budget to samples with semantic drift. \\
\textbf{Diversity ($s_{\mathrm{div}}$).}
Under a fixed IPC budget, selecting only high-confidence samples tends to over-concentrate the subset on a few easy and prototypical modes. To improve coverage, we encourage diversity in the teacher feature space. Let $e_T(x)$ denote the teacher embedding of $x$. We define the diversity of a candidate as its distance to the nearest sample in the currently selected subset:
\begin{equation}
  s_{div}(x)
  = \min_{x' \in \mathcal{S}_c} \left\| e_{{T}}(x) - e_{{T}}(x') \right\|_2^2.
  \label{eq:div}
\end{equation}
Maximizing this term helps prevent budget collapse onto a narrow region of the candidate pool and encourages the selected subset to cover complementary modes. \\
\textbf{Uncertainty ($s_{\mathrm{ent}}$).}
Reliability alone is insufficient, because some candidates may still exhibit ambiguous or unstable semantic predictions. To suppress such samples, we introduce an uncertainty penalty based on the normalized predictive entropy:
\begin{equation}
  s_{ent}(x)
  = - \frac{1}{\log C}\sum_{k=1}^{C} q_{T}^{\tau}(k \mid x)\,
      \log q_{T}^{\tau}(k \mid x),
  \label{eq:entropy}
\end{equation}
where $\tau$ is the temperature scaling parameter. Higher entropy indicates greater semantic ambiguity; therefore, this term is subtracted in the final score to discourage uncertain candidates from consuming scarce budget. \\
\textbf{Greedy Selection.}
We combine the above three criteria into a unified allocation-aware scoring rule. To reduce scale mismatch, each term can be optionally min-max normalized over the pool $\mathcal{P}_c$. The final score for a candidate $x$ is defined as:
\begin{equation}
  s(x) = \alpha\, s_{rel}(x) - \beta\, s_{ent}(x) + \gamma\, s_{div}(x),
  \label{eq:finalscore}
\end{equation}
where $\alpha,\beta,\gamma > 0$ are hyperparameters controlling the trade-off among semantic correctness, ambiguity suppression, and subset coverage. The weights $\alpha,\beta,\gamma$ are set according to the ablation study (Section~\ref{subsubsec:ablation1}).

Given an over-complete class-conditional candidate pool $\mathcal{P}_c$, the goal of Stage I is to allocate the fixed class budget $K$ to a compact subset $\mathcal{S}_c \subset \mathcal{P}_c$ with $|\mathcal{S}_c|=K$ that is most useful for downstream student training. 
we evaluate each candidate $x \in \mathcal{P}_c$ using a composite score that integrates these three criteria. 

We adopt a greedy selection strategy to construct $\mathcal{S}_c$. The subset is initialized with the candidate having the highest reliability score. Then, at each step, we add the candidate that maximizes $s(x)$ and update the diversity term $s_{\mathrm{div}}$ with respect to the current subset. This process continues until $K$ samples are selected.

More recent studies on reliable prediction and trustworthy dataset distillation further suggest that confidence-based signals are closely related to the reliability of recognition models trained under limited or shifted data conditions \cite{10356834}.


\subsection{Soft-Label-Guided Latent Refinement}
\label{latent optimization}
Stage I allocates the fixed IPC budget to more useful seeds from the over-complete candidate pool. However, the selected samples are still only the best available candidates in the current pool, rather than the final optimum for student training. Some selected samples may still suffer from imperfect semantic alignment or local visual artifacts. Therefore, Stage II further refines each selected seed in latent space before constructing the final distilled set.

For each selected seed $(z_i^{\mathrm{init}}, x_i^{\mathrm{init}}) \in \mathcal{S}_c$, we directly use the cached generation latent $z_i^{\mathrm{init}}$ as the optimization variable. Unlike methods that re-encode selected images, our refinement starts from the latent code already used to generate the selected candidate. We optimize only this latent variable while keeping the pretrained generator and VAE decoder frozen. After optimization, the refined latent is denoted as $z_i^{\mathrm{ref}}$, and the corresponding refined image is obtained by $x_i^{\mathrm{ref}}=\mathrm{Dec}(z_i^{\mathrm{ref}})$.
This design preserves the generative prior and avoids modifying the backbone generator during distillation.
We optimize each selected latent using
\begin{equation}
\mathcal{L}_{\mathrm{lat}}(z)=\mathcal{L}_{\mathrm{LDM}}(z)+ w_i\,\mathcal{L}_{\mathrm{KL}}(z),
\label{eq:lat_loss}
\end{equation}
where $\mathcal{L}_{\mathrm{LDM}}$ is the denoising objective reused from Section~\ref{subsec:LDM} as an on-manifold regularizer, and $\mathcal{L}_{\mathrm{KL}}$ aligns the refined sample with the cached teacher soft target associated with the selected seed. The effect of soft-label supervision is analyzed in Section~\ref{subsubsec:ablation2}. \\
\textbf{Sample-adaptive weighting.}
Let $p_i = q_T^\tau(c \mid x_i^{\mathrm{init}})$ denote the cached target-class confidence of the selected seed. We use
\begin{equation}
   w_i = \lambda \cdot \bigl( w_{\min} + (1-p_i)(w_{\max}-w_{\min}) \bigr),
\label{eq:weight}
\end{equation}
so that less confident seeds receive stronger semantic alignment, while highly reliable seeds are refined more conservatively.
For efficiency, we adopt a late-stage schedule by disabling $\mathcal{L}_{\mathrm{KL}}$ during the first $L-R$ refinement iterations and enabling it only in the final $R$ steps. This allows the latent to first adapt to the diffusion prior and then receive stronger semantic correction near the end of refinement.

After refinement, each refined image $x_i^{\mathrm{ref}}=\mathrm{Dec}(z_i^{\mathrm{ref}})$ is paired with its cached teacher soft target $q_i^{\mathrm{ref}}$ and added to the final distilled set. The student training protocol on these refined pairs is summarized in Section~\ref{algorithmic flow}.

\subsection{Algorithmic Flow and Student Training}
\label{algorithmic flow}
\begin{algorithm}[!htbp]
\caption{Pool-Select-Refine}
\label{alg:main}
\begin{algorithmic}[1]
\Ensure Distilled set $\mathcal{D}^\star$
\Require Teacher $T$, generator $G$ (decoder $\mathrm{Dec}$), student $S$;
classes $C$; IPC $K$; pool size $M$; temperature $\tau$;
weights $(\alpha,\beta,\gamma)$; refine params $(\lambda,w_{\min},w_{\max},\eta,L,R)$.
\State $\mathcal{D}^\star \gets \emptyset$
\For{$c = 1$ to $C$}
  \State Sample $\{z_i^{\mathrm{init}}\}_{i=1}^{M} \sim p(z \mid c)$
  \State Decode candidates $x_i^{\mathrm{init}} \gets \mathrm{Dec}(z_i^{\mathrm{init}})$
  \State Query teacher soft labels $q_i \gets q_{T}^\tau(\cdot \mid x_i^{\mathrm{init}})$
  \State Extract teacher embeddings $e_i \gets e_{T}(x_i^{\mathrm{init}})$
  \State Keep only samples predicted as class $c$ by $T$
  \State Initialize selected set $\mathcal{S}_c \gets \emptyset$
  \For{$t = 1$ to $K$}
    \State Pick next index $i^\star \in \mathcal{I}_c \setminus \mathcal{S}_c$
    \State Update selection score with Equation~\ref{eq:finalscore}
    \State $\mathcal{S}_c \gets \mathcal{S}_c \cup \{i^\star\}$
  \EndFor
  \For{each $i \in \mathcal{S}_c$}
    \State Cache teacher target $q_i^{\mathrm{ref}} \gets q_i$
    \State Initialize latent $z \gets z_i^{\mathrm{init}}$
    \State Update adaptive weight $w_i$ with Equation~\ref{eq:weight}
    \For{$\ell = 1$ to $L$}
      \State Update objective with Equation~\ref{eq:lat_loss}
      \State \hspace{1.2em} Apply late-stage KL only if $\ell > L-R$
      \State $z \gets z - \eta \nabla_z \mathcal{L}_{\mathrm{lat}}$
    \EndFor
    \State Set $z_i^{\mathrm{ref}} \gets z$
    \State Add refined pair $(\mathrm{Dec}(z_i^{\mathrm{ref}}), q_i^{\mathrm{ref}})$ into $\mathcal{D}^\star$
  \EndFor
\EndFor
\State Train student $S$ on $\mathcal{D}^\star$ with Equation~\ref{eq:kd_cached}
\State \Return $\mathcal{D}^\star$
\end{algorithmic}
\end{algorithm}

Algoruthmic~\ref{alg:main} summarizes the full ``\textit{Pool-Select-Refine}'' pipeline. For each class, we first construct an over-complete candidate pool, allocate the fixed IPC budget through Stage I selection, and then refine the selected seeds in latent space via Stage II before forming the final distilled set.
After Stage II, the final distilled set is represented as pairs $(x_i^{\mathrm{ref}}, q_i^{\mathrm{ref}})$, where $x_i^{\mathrm{ref}}=\mathrm{Dec}(z_i^{\mathrm{ref}})$ is the refined distilled image and $q_i^{\mathrm{ref}}$ is the cached teacher soft target inherited from the selected seed.
The student is then trained on $\mathcal{D}^\star$ using temperature-scaled knowledge distillation:
\begin{equation}
\mathcal{L}_{\mathrm{KD}}=
\frac{\tau^{2}}{|\mathcal{D}^\star|}
\sum_{(x^{\mathrm{ref}},q^{\mathrm{ref}})\in\mathcal{D}^\star}
\mathrm{KL}\!\left(q^{\mathrm{ref}}\ \big\|\ q_{S}^{\tau}(\cdot \mid x^{\mathrm{ref}})\right).
\label{eq:kd_cached}
\end{equation}
Unless otherwise stated, this KD objective is used as the default student training protocol throughout the paper. For completeness, we also compare against hard-label supervision in the Section~\ref{subsubsec:ablation3}. \\

\section{Experiments}
\subsection{Setting}
\label{subsec:setting}

\begin{table*}[t]
\centering
\setlength{\tabcolsep}{4pt}
\renewcommand{\arraystretch}{1.15}
\caption{\textbf{Quantitative results.} Comparison of classification accuracy (\%) under different IPC settings on ImageWoof subsets. MInimax-Ours consistently outperforms baselines under all IPC levels, DiT-Ours consistently improves over DiT and is competitive with prior baselines (with some settings where DM remains stronger). Notably, under extremely low IPC (e.g., IPC=1), our distilled sets maintain strong generalization, substantially improves over baselines under extreme low IPC. N/A means the referenced method does not report results for the corresponding IPC/model configuration, so a direct comparison is unavailable.}
\resizebox{\textwidth}{!}{
\begin{tabular}{llcccccccccc}
\toprule
\multicolumn{1}{c}{\textbf{IPC (Ratio)}} & \multicolumn{1}{c}{\textbf{Test Model}}
& \textbf{Random} & \textbf{DM} \cite{690345d30c9f4e6dbeef4ad9210bee62} & \textbf{IDC-1} \cite{kim2022dataset} & \textbf{GLaD} \cite{cazenavette2023glad} & \textbf{RDED} \cite{Sun2024RDED}
& \textbf{DiT} & \textbf{Minimax} \cite{Gu_2024_CVPR}
& \multicolumn{1}{c}{\underline{\textbf{DiT-Ours}}}
& \multicolumn{1}{c}{\underline{\textbf{Minimax-Ours}}}
& \textbf{Full} \\
\midrule
\multirow{3}{*}{\textbf{1} (0.08\%)}
& ConvNet-6   & $14.2\pm0.9$ & $21.1\pm0.5$ & N/A & N/A & $18.5\pm0.9$ & $12.7\pm0.6$ & $15.2\pm0.6$ & $18.9\pm0.4$ & $\mathbf{21.5\pm0.9}$ & $86.4\pm0.2$ \\
& ResNetAP-10 & $17.8\pm2.4$ & N/A          & N/A & N/A & N/A           & $18.0\pm1.3$ & $18.9\pm2.4$ & $22.8\pm1.2$ & $\mathbf{23.1\pm1.5}$ & $87.5\pm0.5$ \\
& ResNet-18   & $13.5\pm0.4$ & N/A          & N/A & N/A & $20.8\pm1.2$ & $15.3\pm0.7$ & $14.6\pm0.6$ & $20.7\pm0.5$ & $\mathbf{21.9\pm0.9}$ & $89.3\pm1.2$ \\
\midrule
\multirow{3}{*}{\textbf{10} (0.4\%)}
& ConvNet-6   & $24.3\pm1.1$ & $26.9\pm1.2$ & $33.3\pm1.1$ & $33.8\pm0.9$ & $40.6\pm2.0$ & $34.2\pm1.1$ & $37.0\pm1.0$ & $40.6\pm0.8$ & $\mathbf{40.8\pm1.3}$ & $86.4\pm0.2$ \\
& ResNetAP-10 & $29.4\pm0.8$ & $30.3\pm1.2$ & $39.1\pm0.5$ & $32.9\pm0.9$ & N/A           & $34.7\pm0.5$ & $39.2\pm1.3$ & $\mathbf{42.0\pm0.7}$ & $40.8\pm0.4$ & $87.5\pm0.5$ \\
& ResNet-18   & $27.7\pm0.9$ & $33.4\pm0.7$ & $37.3\pm0.2$ & $31.7\pm0.8$ & $38.5\pm2.1$ & $34.7\pm0.4$ & $37.6\pm0.9$ & $\mathbf{41.2\pm0.3}$ & $39.3\pm0.9$ & $89.3\pm1.2$ \\
\midrule
\multirow{3}{*}{\textbf{20} (1.6\%)}
& ConvNet-6   & $29.1\pm0.7$ & $29.9\pm1.0$ & $35.5\pm0.8$ & N/A & N/A & $36.1\pm0.8$ & $37.6\pm0.2$ & $45.6\pm1.1$ & $\mathbf{48.4\pm0.8}$ & $86.4\pm0.2$ \\
& ResNetAP-10 & $32.7\pm0.4$ & $35.2\pm0.6$ & $43.3\pm0.3$ & N/A & N/A & $41.1\pm0.8$ & $45.8\pm0.5$ & $48.2\pm1.7$ & $\mathbf{50.4\pm0.7}$ & $87.5\pm0.5$ \\
& ResNet-18   & $29.7\pm0.5$ & $29.8\pm1.7$ & $38.6\pm0.2$ & N/A & N/A & $40.5\pm0.5$ & $42.5\pm0.6$ & $46.4\pm0.6$ & $\mathbf{47.6\pm1.3}$ & $89.3\pm1.2$ \\
\midrule
\multirow{3}{*}{\textbf{50} (3.8\%)}
& ConvNet-6   & $41.3\pm0.6$ & $44.4\pm1.0$ & $43.9\pm1.2$ & N/A & $61.5\pm0.3$ & $46.5\pm0.8$ & $53.9\pm0.6$ & $56.6\pm0.7$ & $\mathbf{62.9\pm1.0}$ & $86.4\pm0.2$ \\
& ResNetAP-10 & $47.2\pm1.3$ & $47.1\pm1.1$ & $48.3\pm1.0$ & N/A & N/A           & $49.3\pm0.2$ & $56.3\pm1.0$ & $58.8\pm0.6$ & $\mathbf{63.8\pm1.6}$ & $87.5\pm0.5$ \\
& ResNet-18   & $47.9\pm1.8$ & $46.2\pm0.6$ & $48.3\pm0.8$ & N/A & N/A           & $50.1\pm0.5$ & $57.1\pm0.6$ & $57.3\pm1.8$ & $\mathbf{62.7\pm0.4}$ & $89.3\pm1.2$ \\
\midrule
\multirow{3}{*}{\textbf{100} (7.7\%)}
& ConvNet-6   & $52.2\pm0.4$ & $55.0\pm1.3$ & $53.2\pm0.9$ & N/A & N/A           & $53.4\pm0.3$ & $61.1\pm0.7$ & $60.9\pm0.4$ & $\mathbf{68.7\pm1.2}$ & $86.4\pm0.2$ \\
& ResNetAP-10 & $59.4\pm1.0$ & $56.4\pm0.8$ & $56.1\pm0.9$ & N/A & N/A           & $58.3\pm0.8$ & $64.5\pm0.2$ & $63.6\pm0.6$ & $\mathbf{72.6\pm1.1}$ & $87.5\pm0.5$ \\
& ResNet-18   & $61.5\pm1.3$ & $60.2\pm1.0$ & $58.3\pm1.2$ & N/A & N/A           & $58.9\pm1.3$ & $65.7\pm0.4$ & $62.4\pm1.2$ & $\mathbf{69.8\pm1.4}$ & $89.3\pm1.2$ \\
\bottomrule
\end{tabular}}
\label{tab:experment1}
\end{table*}

We evaluate our method on multiple distillation settings with varying dataset scales and architectures. Table~\ref{tab:experment1} reports results on ImageWoof using three student architectures: ConvNet-6, ResNet-AP-10, and ResNet-18, following prior work~\cite{gidaris2018dynamic,He2016ResNet}. We test five IPC levels, $K \in \{1,10,20,50,100\}$, covering both extremely low-data and relatively relaxed regimes. The teacher is a pretrained ConvNeXt-B \cite{Liu2022ConvNeXt}, and the evaluation metric is top-1 accuracy reported as mean $\pm$ standard deviation over three runs. We use the released generative/distillation pipelines of DiT and Minimax to synthesize class-conditional candidate pools; “Minimax” in the tables refers to the corresponding baseline distillation pipeline. Unless otherwise stated, we set the sampling steps = 50, candidate pool size to $M\ge2K$, use temperature $\tau=1.5$, optimize each selected latent for $L=100$ iterations with learning rate $\eta=0.01$, and set $\lambda=0.5$, $w_{\min}=0.1$, $w_{\max}=2.0$, and $R=4$ for Stage II refinement.

For Stage I selection, we use an IPC-dependent preset for $(\alpha,\beta,\gamma)$: $(3.0,1.0,0.5)$ for $K \le 10$, $(1.0,1.0,1.0)$ for $K=20$, and $(0.5,1.0,3.0)$ for $K \ge 50$. Unless otherwise stated, all student models are trained from scratch on the refined distilled pairs $(x,q^{ref})$ using the temperature-scaled KD objective in Section~\ref{algorithmic flow}.

\subsection{Results on ImageNet Subsets}
\label{subsec:result}

\begin{table}[t]
\small
\centering
\footnotesize
\setlength{\tabcolsep}{2.5pt}
\renewcommand{\arraystretch}{0.60}
\caption{\textbf{Comparison of test accuracy (\%) on ImageNette/IDC under different IPC settings.} Minimax-Ours consistently achieves the best performance; DiT-Ours improves upon DiT and is competitive across settings.}
\resizebox{0.92\columnwidth}{!}{
\begin{tabular}{c c c c c c c}
\toprule
\multicolumn{1}{l}{} & \textbf{IPC} & \textbf{Random} & \textbf{DiT} & \textbf{Minimax}
& \underline{\textbf{DiT-Ours}} & \underline{\textbf{Minimax-Ours}} \\
\midrule
\multirow{3}{*}{\rotatebox{90}{\textbf{Nette}}}
& 10 & $54.2\pm1.6$ & $59.1\pm0.7$ & $62.0\pm0.2$ & $60.2\pm1.0$ & $\mathbf{63.1\pm1.9}$ \\
& 20 & $63.5\pm0.5$ & $64.8\pm1.2$ & $66.8\pm0.4$ & $66.7\pm0.8$ & $\mathbf{68.3\pm0.9}$ \\
& 50 & $76.1\pm1.1$ & $73.3\pm0.9$ & $76.6\pm0.2$ & $78.7\pm1.1$ & $\mathbf{81.1\pm1.7}$ \\
\midrule
\multirow{3}{*}{\rotatebox{90}{\textbf{IDC}}}
& 10 & $48.1\pm0.8$ & $54.1\pm0.4$ & $53.1\pm0.2$ & $\mathbf{56.6\pm0.5}$ & $54.8\pm1.5$ \\
& 20 & $52.5\pm0.9$ & $58.9\pm0.2$ & $59.0\pm0.4$ & $62.7\pm0.6$ & $\mathbf{65.3\pm1.1}$ \\
& 50 & $68.1\pm0.7$ & $64.3\pm0.6$ & $69.6\pm0.2$ & $67.8\pm1.6$ & $\mathbf{70.8\pm1.4}$ \\
\bottomrule
\end{tabular}}
\label{tab:experment2}
\vspace{-4pt}
\end{table}

Table~\ref{tab:experment1} reports the main results on ImageWoof across five IPC settings and three student architectures. Our method consistently improves over its corresponding diffusion-based generators under the same generative prior. In particular, Minimax-Ours achieves consistent gains over Minimax across all IPC levels and evaluation models, while DiT-Ours also stably improves upon DiT and remains competitive with strong prior baselines. These results support our central claim: under a fixed IPC budget, explicitly decoupling candidate generation from IPC-constrained curation and subsequent refinement leads to more effective use of the distilled budget than directly adopting a rigid ``\textit{Generate-and-Use}'' pipeline.

Table~\ref{tab:experment2} further evaluates the proposed framework on two fine-grained ImageNet subsets, ImageNette \cite{Deng2009ImageNet,Howard2019ImageNette} and ImageIDC \cite{Deng2009ImageNet,Zhao2021DSA}, where classes exhibit higher intra-class similarity and are therefore more challenging for compact distillation. The same trend remains clear. Minimax-Ours consistently achieves the strongest performance, and DiT-Ours generally improves upon DiT while remaining competitive across settings. These results indicate that the proposed two-stage design is not limited to coarse-grained scenarios: even in fine-grained regimes, explicitly introducing a curation stage before latent refinement helps allocate limited budget to more useful synthetic samples.

\subsection{Ablation}
\label{subsec:ablation}

\subsubsection{Complementarity of selection signals}
\label{subsubsec:ablation1}

\begin{figure}[t]
    \centering
    \includegraphics[width=1.0\linewidth]{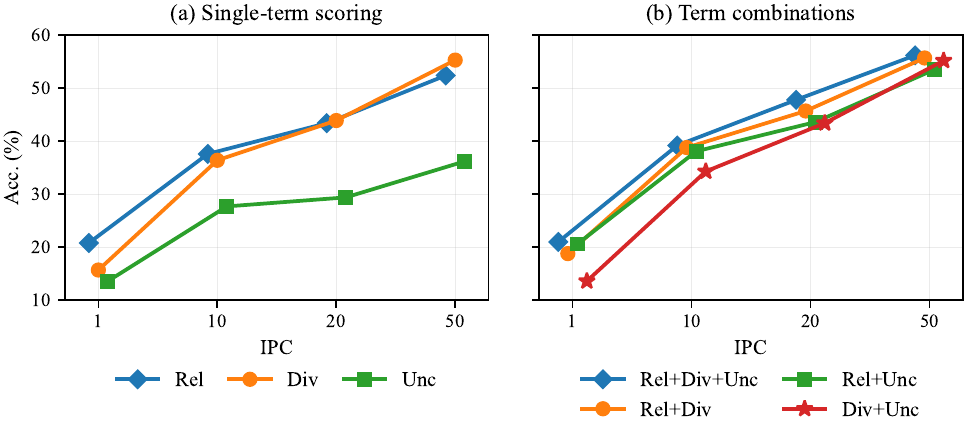}
    \caption{\textbf{Ablation of selection signals under a DiT generative prior.} (a) Single-term scoring using reliability, diversity, or uncertainty. (b) Pairwise and full combinations. Pairwise scoring consistently improves over single-term variants, and using all three signals achieves the best or tied-best accuracy across IPC settings, indicating complementary roles in subset curation. Markers are slightly horizontally shifted for better readability.}
    \label{fig:ablation1}
\end{figure}

To investigate the impact of each scoring component in our selection strategy, we conduct an ablation study in which only one of the three terms—reliability, uncertainty, or diversity—is activated at a time. As shown in Figure~\ref{fig:ablation1}(a), the Rel (reliability term) achieves the best performance in low-budget regimes (IPC $\le 10$), while the Div (diversity term) gradually outperforms the others as IPC increases. The Unc (uncertainty term), when used alone, is less effective, especially in small-IPC settings.

Importantly, Figure~\ref{fig:ablation1}(b) demonstrates synergy: any pairwise combination consistently outperforms the corresponding single-term variants, and combining all three signals yields the best (or tied-best) performance across IPC settings. This supports our design choice in Section~\ref{sample selection} to score candidates jointly by reliability, diversity, and uncertainty rather than relying on a single heuristic.

Based on this, we use a simple preset schedule (chosen once from Figure~\ref{fig:ablation1} and fixed for all datasets/backbones). For low-data regimes ($K\le 10$), we set $(\alpha,\beta,\gamma)=(3.0,1.0,0.5)$ to prioritize semantic correctness. For large budgets ($K\ge 50$), we set $(\alpha,\beta,\gamma)=(0.5,1.0,3.0)$ to emphasize coverage. For intermediate budgets (e.g., $K=20$), we use the balanced setting $(\alpha,\beta,\gamma)=(1.0,1.0,1.0)$. All weights are shared across classes within the same generator.

\subsubsection{Effectiveness of Soft-Label KL in Latent Optimization}
\label{subsubsec:ablation2}
\begin{figure}[t]
    \centering
    \includegraphics[width=1.0\linewidth]{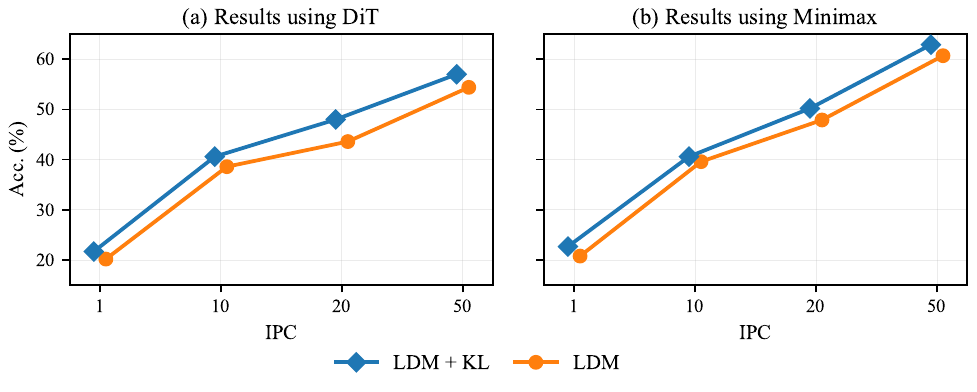}
    \caption{\textbf{Effect of soft-label guidance in latent-space optimization.} We compare the final classification accuracy between latent optimization with (LDM + KL) and without KL. Results on both DiT-based (a) and Minimax-based (b) pipelines demonstrate consistent performance improvement from soft-label supervision in the latent space. Markers are slightly horizontally shifted for better readability.}
    \label{fig:ablation2}
\end{figure}

We further validate the role of the soft-label KL divergence term in the latent optimization stage. As shown in Figure~\ref{fig:ablation2}, adding the KL loss on top of the LDM \cite{Rombach2022LDM} consistency significantly improves performance across all IPC settings. The improvement is particularly noticeable in medium and high IPC ranges, suggesting that aligning latent samples with teacher-predicted soft targets helps guide the optimization toward more class-discriminative and semantically aligned representations. Based on these observations, we include the KL term in the refinement objective for all experiments, using the late-stage schedule described in Section~\ref{latent optimization}.

\subsubsection{Soft Labels v.s. Hard Labels in Student Training}
\label{subsubsec:ablation3}
\begin{figure}[t]
    \centering
    \includegraphics[width=1.0\linewidth]{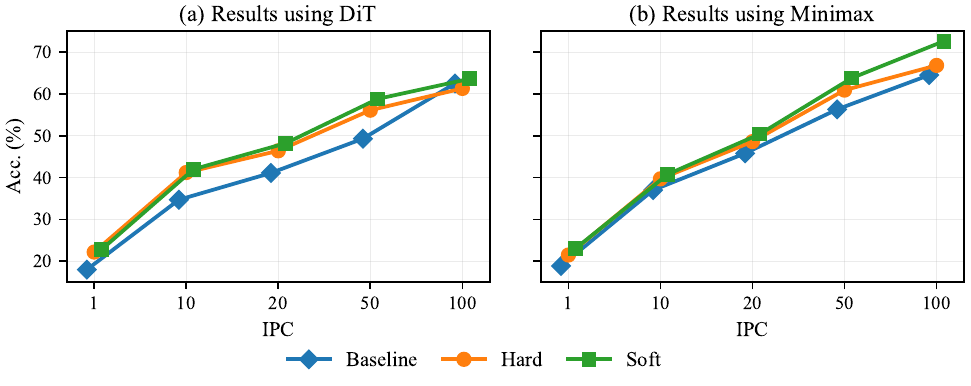}
    \caption{\textbf{Comparison of soft-label and hard-label supervision when training models on distilled datasets.} We report classification accuracy using three label settings—baseline (from original synthetic data), hard labels and soft labels. Across both DiT (a) and Minimax (b) pipelines, soft-label training consistently yields superior accuracy, especially under low IPC conditions. Markers are slightly horizontally shifted for better readability.}
    \label{fig:ablation3}
\end{figure}

We compare the effect of using soft labels versus hard labels when training the student network on the selected synthetic samples. Figure~\ref{fig:ablation3} summarizes the accuracy under both settings using DiT and Minimax \cite{Gu_2024_CVPR} generators, respectively. Across all IPC levels, soft-label supervision yields equal or better accuracy than hard labels, with consistent gains in the medium to high IPC range. This supports the intuition that soft targets provide richer inter-class relations and prevent over-confident predictions during low-data training. We therefore adopt soft labels by default in the student distillation phase.

\subsubsection{Complementarity of selection signals in Minimax}
\label{subsubsec:minimax}
\begin{figure}[t]
    \centering
    \includegraphics[width=1.0\linewidth]{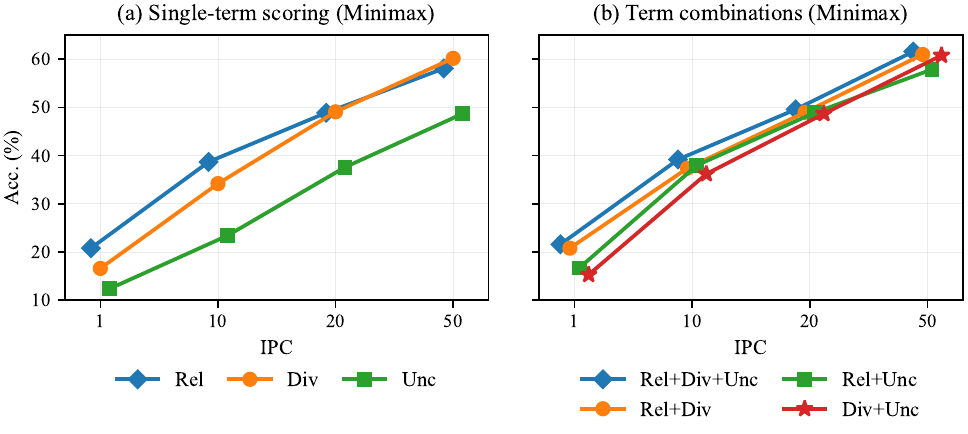}
    \caption{\textbf{Complementarity of selection signals under a Minimax generative prior.} (a) Single-term scoring using reliability, diversity or uncertainty. (b) Term combinations. Pairwise combinations generally outperform single-term variants, and combining all three signals yields the best or tied-best performance across IPC settings, indicating complementary roles in subset curation. Markers are slightly horizontally shifted for better readability.}
    \label{fig:appendix0}
\end{figure}

\begin{table}[t]
\centering
\caption{\textbf{Performance comparison on CIFAR-10, CIFAR-100, and Tiny-ImageNet under IPC = 1, 10, and 50. }
Our method (DiT-Ours) achieves strong performance across datasets, outperforming DM/DC on CIFAR-100 and Tiny-ImageNet across IPC settings, and showing clear gains on CIFAR-10 at higher IPC(e.g., IPC=50). N/A means the referenced method does not report results for the corresponding IPC/model configuration, so a direct comparison is unavailable.}
\label{tab:appendix2}
\resizebox{\textwidth}{!}{
\begin{tabular}{lccc ccc ccc}
\toprule
& \multicolumn{3}{c}{\textbf{CIFAR-10}} 
& \multicolumn{3}{c}{\textbf{CIFAR-100}}
& \multicolumn{3}{c}{\textbf{Tiny-ImageNet}} \\
\cmidrule(lr){2-4} \cmidrule(lr){5-7} \cmidrule(lr){8-10}
\textbf{IPC}  
& \textbf{1} & \textbf{10} & \textbf{50}
& \textbf{1} & \textbf{10} & \textbf{50}
& \textbf{1} & \textbf{10} & \textbf{50} \\
\midrule
DM    
& $\mathbf{26.0\!\pm\!0.8}$ & $\mathbf{48.9\!\pm\!0.6}$ & $63.0\!\pm\!0.4$
& $11.4\!\pm\!0.3$ & $29.7\!\pm\!0.3$ & $43.6\!\pm\!0.4$
& $3.9\!\pm\!0.2$  & $12.9\!\pm\!0.4$ & $24.1\!\pm\!0.3$ \\
DC  
& $28.3\!\pm\!0.5$ & $44.9\!\pm\!0.5$ & $53.9\!\pm\!0.5$
& $12.8\!\pm\!0.3$ & $25.2\!\pm\!0.3$ & N/A
& N/A & N/A & N/A \\
\underline{DiT-Ours} 
& $22.3\!\pm\!0.8$ & $45.7\!\pm\!0.6$ & $\mathbf{66.8\!\pm\!1.1}$
& $\mathbf{16.4\!\pm\!0.9}$ & $\mathbf{46.2\!\pm\!0.7}$ & $\mathbf{56.8\!\pm\!0.5}$
& $\mathbf{15.6\!\pm\!0.2}$ & $\mathbf{35.9\!\pm\!0.6}$ & $\mathbf{42.8\!\pm\!1.2}$ \\
\bottomrule
\end{tabular}
}
\end{table}

To verify that the observed complementarity is not specific to a particular generator, we repeat the scoring ablation using Minimax Diffusion as the generative prior, following the same protocol as in Section~\ref{subsubsec:ablation1}. Figure~\ref{fig:appendix0}(a) shows that single-term scoring is insufficient: reliability tends to be more effective in low-IPC regimes, while diversity becomes increasingly important as IPC grows; using uncertainty alone remains less competitive. More importantly, Figure~\ref{fig:appendix0}(b) indicates clear synergy—pairwise combinations generally improve over single-term variants, and the full combination (Rel+Div+Unc) achieves the best (or near-best) accuracy across IPC settings. These results mirror the DiT-based observations in Section~\ref{subsubsec:ablation1}, supporting  the conclusion that our composite scoring captures complementary cues and transfers across diffusion-based generators.

\subsubsection{Distill other datasets}
\label{subsubsec:distill}
To assess the generality of our method beyond ImageNet subsets, we further evaluate it on CIFAR-10, CIFAR-100 \cite{Krizhevsky2009CIFAR}, and Tiny-ImageNet \cite{Le2015TinyImageNet}. As shown in Table~\ref{tab:appendix2}, our method outperforms DM \cite{690345d30c9f4e6dbeef4ad9210bee62} and DC \cite{DBLP:journals/corr/abs-2006-05929} on CIFAR-100 and Tiny-ImageNet across IPC settings. On CIFAR-10, our method becomes advantageous at higher IPC (e.g., IPC=50), while remaining competitive at IPC=1/10.

\subsubsection{Combined with other methods}
\label{subsubsec:combined}
To verify the compatibility and generality of our strategy, we integrate our selection and optimization approach with two strong existing distillation frameworks: DiT+IGD \cite{chen2025influenceguided} and MTT+GLaD \cite{cazenavette2023glad}. As shown in Table~\ref{tab:appendix3}, our method consistently improves classification accuracy across different IPC levels and distillation backbones. For instance, on DiT+IGD, the gain is particularly significant at IPC=100 (+6.7\%), and for \\
\begin{wraptable}{r}{0.46\textwidth}
\vspace{-16pt}
\centering
\footnotesize
\setlength{\tabcolsep}{8pt}
\renewcommand{\arraystretch}{0.60}
\caption{\textbf{Integration with existing methods.}}
\label{tab:appendix3}
\begin{tabular}{lccc}
\toprule
\textbf{Method} & \textbf{10} & \textbf{50} & \textbf{100} \\
\midrule
DiT+IGD & 41.0 & 62.7 & 69.7 \\
\underline{+ Ours} & \textbf{42.3} & \textbf{65.6} & \textbf{76.4} \\
\midrule
MTT+GLaD & 17.1 & N/A & N/A \\
\underline{+ Ours} & \textbf{22.6} & \textbf{30.8} & \textbf{55.1} \\
\bottomrule
\end{tabular}
\vspace{-8pt}
\end{wraptable}
MTT+GLaD, our strategy boosts the previously underperforming IPC=10 case from 17.1\% to 22.6\%. These results confirm that our strategy is not only plug-and-play but also broadly effective, suggesting the potential for future unification across generative and trajectory-based distillation paradigms.

\subsection{Analysis of Hyperparameters}
\label{analysis of hyperparameter}
\begin{figure}[t]
    \centering
    \includegraphics[width=1.0\linewidth]{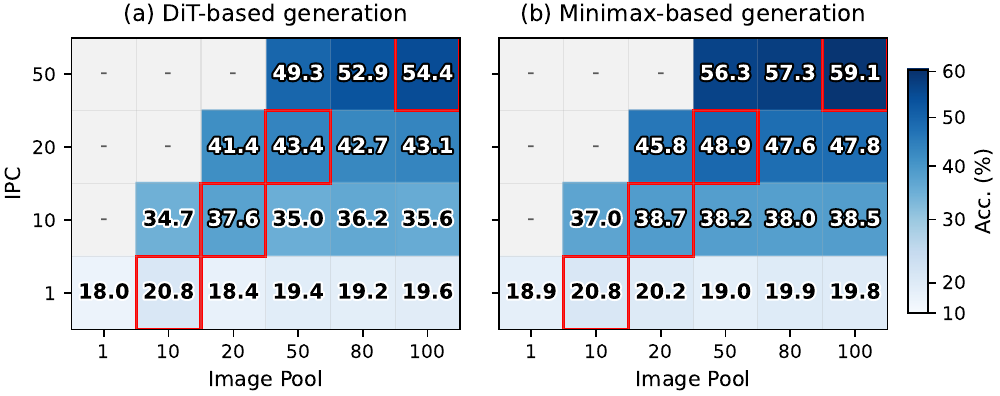}
    \caption{\textbf{Heatmaps of classification accuracy over combinations of per-class image pool sizes and IPC.} Each cell shows the classification accuracy when $K$ samples from a pool of size $M$ per class.Red boxes indicate the best configuration for each target IPC.}
    \label{fig:intro4}
\end{figure}

A critical hyperparameter is the size of the candidate pool $M$. Figure~\ref{fig:intro4} presents the accuracy heatmaps for varying pool sizes versus target IPC ($K$). We observe a clear linear scaling law: the optimal pool size is approximately $2 \times K$ (twice the target IPC). \\
\textbf{Too small($M \approx K$)}. The selection space is overly constrained, degenerating into direct generation. \\
\textbf{Too large($M \gg 2K$)}. Performance tends toward saturation or shows slight decline, likely due to the inclusion of outliers in the candidate pool. This finding provides practical guidance for future research: merely doubling the budget yields selection advantages while keeping computational overhead within manageable limits.

\subsection{Visualization}
\label{visualization}

\begin{figure*}[t]
    \centering
    \includegraphics[width=0.9\linewidth]{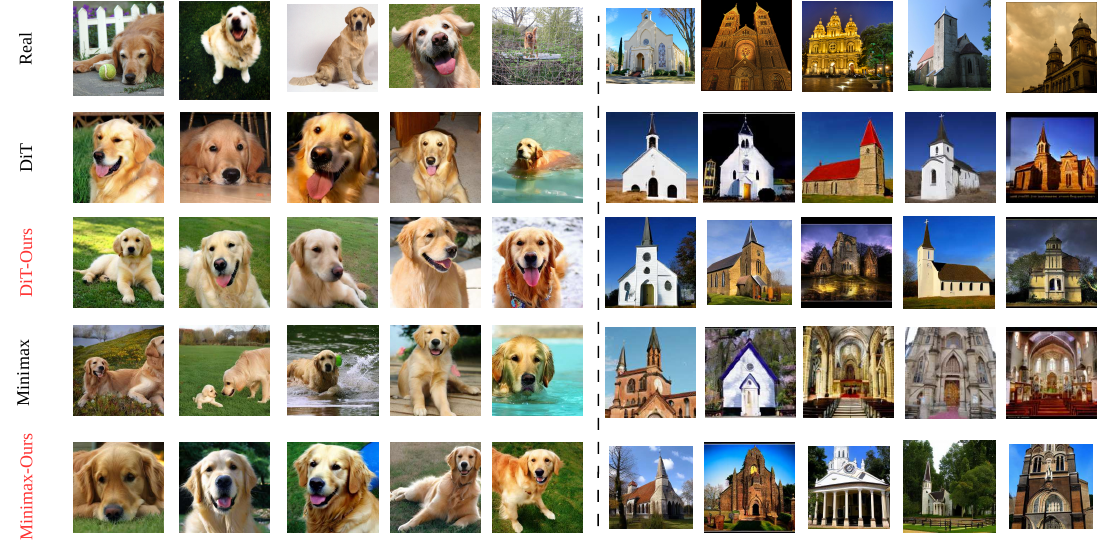}
    \caption{\textbf{Qualitative visualization of distilled samples.} We compare images from the Real dataset (Top row) against synthetic samples generated by standard baselines (DiT, Minimax) and our proposed method (DiT-Ours, Minimax-Ours). Columns exhibit samples from the \textit{Golden Retriever} and \textit{Church} classes. Compared to the direct generation baselines, which occasionally yield ambiguous backgrounds or artifacts, our method synthesizes images with higher semantic fidelity and clearer structural details (e.g., distinct facial features of dogs and architectural geometry of churches), demonstrating the efficacy of our diversity-aware selection and latent refinement.}
    \label{fig:visualization}
\end{figure*}

To intuitively assess the quality of the distilled dataset, we visualize the synthesized samples in Figure~\ref{fig:visualization}. We compare the images generated by our proposed method (applied to both DiT and Minimax backbones) against the original images and standard Minimax \cite{Gu_2024_CVPR} baseline. \\
\textbf{Visual Fidelity and Reliability}. As observed in the ``Golden Retriever'' class, the baseline (Minimax) occasionally generates samples with blurred backgrounds or insufficiently defined facial features. In contrast,  samples generated by Minimax-Ours exhibit greater visual \textit{typicality}—capturing key semantic features with exceptional sharpness, such as fur texture and nasal structure. This qualitative leap validates our reliability-based selection strategy, which effectively filters low-quality or noisy modes within the candidate pool. \\
\textbf{Semantic Alignment and Diversity}. Within the \textit{church} class, real images exhibit significant variations in lighting and architectural structures. Our approach (DiT-Ours) successfully synthesizes diverse architectural styles—such as varying spire shapes and viewpoints—while maintaining structural consistency. Unlike baseline models prone to geometric distortions, our samples preserve sharp edges and accurate perspective effects. This structural integrity stems primarily from our soft-label-guided latent optimization technique—which forces the latent encoder to closely align with the teacher's precise semantic understanding, correcting latent artifacts before final decoding.

These visual results confirm that our performance gains (Tabel~\ref{tab:experment1},Table~\ref{tab:experment2}) are not merely numerical artifacts but stem from a tangible improvement in the semantic density and visual quality of the synthetic data.

\section{Conclusion}
In this paper, we studied diffusion-based dataset distillation from a budget-allocation perspective. We observed that the common ``\textit{Generate-and-Use}'' paradigm directly treats generated samples as the final distilled set, which may waste limited IPC slots on redundant, weakly informative, or semantically unstable candidates. To address this issue, we proposed ``\textit{Pool-Select-Refine}'', a two-stage framework that first constructs an over-complete class-conditional candidate pool, selects a compact subset using teacher-derived reliability, diversity, and uncertainty, and then refines the selected latent codes with soft-label guidance while keeping the pretrained generative prior frozen.

The strength of the proposed framework lies in its explicit control over how each limited IPC slot is allocated. Experiments on ImageNet subsets and fine-grained benchmarks show that our method consistently improves the corresponding diffusion-based baselines, especially in low-budget settings where each selected sample has a large influence on student training. The ablation studies also show that the three selection signals are complementary and that latent refinement further improves the selected candidates. More broadly, this work suggests that generative DD should not only focus on producing realistic images, but also on selecting and refining samples that provide reliable and diverse supervision for compact recognition.

There are still several limitations. The selection process depends on a fixed teacher model, so biased or poorly calibrated teacher predictions may affect the selected subset. The framework also introduces additional computation for pool generation, teacher evaluation, feature extraction, and latent refinement, although it avoids retraining the generator. In addition, the current scoring weights are empirically chosen and may need adjustment across datasets and IPC budgets. Future work will explore adaptive weighting strategies, more efficient pool construction and refinement schedules, robustness to teacher bias, and extensions to other recognition tasks such as detection, segmentation, multimodal recognition, and domain-shifted scenarios.

\section*{CRediT authorship contribution statement}
\textbf{Wenmin Li}: Conceptualization, Methodology, Software, Validation, Formal analysis, Investigation, Writing -- original draft, Writing -- review \& editing, Visualization.
\textbf{Zhongkai Zhao}: Methodology, Supervision, Writing -- review \& editing.
\textbf{Shunsuke Sakai}: Methodology, Supervision, Writing -- review \& editing.
\textbf{Tatsuhito Hasegawa}: Resources, Funding acquisition, Supervision, Writing -- review \& editing.

\section*{Declaration of competing interest} The authors declare that they have no known competing financial interests or personal relationships that could have appeared to influence the work reported in this paper.

\section*{Data availability}
Data will be made available on request.

\section*{Acknowledgements} This work was supported in part by the Japan Society for the Promotion of Science (JSPS) KAKENHI Grant-in-Aid for Scientific Research A (25H01110), B (26K02879), C (23K11164), and the Grant by Marine Informatics Research Institute.

\section*{Declaration of generative AI and AI-assisted technologies in the writing process}
During the preparation of this work the authors used ChatGPT by OpenAI in order to assist with language polishing, grammar checking, formatting, and improving the clarity of the manuscript and submission materials. After using this tool/service, the authors reviewed and edited the content as needed and take full responsibility for the content of the publication. The tool was not used to generate research data, perform experiments, conduct analyses, or draw scientific conclusions.

\bibliographystyle{unsrt}

\bibliography{mybib}


\end{document}